\documentclass[envcountsame,runningheads]{llncs}
\usepackage{xspace}
\usepackage{amsmath}
\usepackage{amssymb}
\usepackage{graphicx}
\newcommand{\ct}{\textsf{c}\texttt{\char13}\textsf{t}\xspace}
\newcommand{\myurl}[1]{\url{\texttt{#1}}\xspace}
\newcommand{\IN}{\mathbb{N}}
\newcommand{\calO}{\mathcal{O}}
\newcommand{\person}[1]{\textsc{#1}}
\newcommand{\aname}[1]{\textsf{#1}}
\newcommand{\mycite}[2]{{\rm\cite[\textsc{#1}]{#2}}}
\newcommand{\COMMENTED}[1]{}

\spnewtheorem{observation}[theorem]{Observation}{\bfseries}{\itshape}
\spnewtheorem{fact}[theorem]{Fact}{\bfseries}{\itshape}
\spnewtheorem{myclaim}[theorem]{Claim}{\bfseries}{\itshape}
\spnewtheorem{scholium}[theorem]{Scholium}{\bfseries}{\itshape}
\spnewtheorem{myexample}[theorem]{Example}{\bfseries}{\itshape}
\spnewtheorem{myremark}[theorem]{Remark}{\bfseries}{\itshape}

\begin{document}
\setcounter{secnumdepth}{3}
\setcounter{tocdepth}{3}
\title{Variations of the Turing Test\newline 
in the Age of Internet and Virtual Reality}
\titlerunning{Variations of the Turing Test 
in the Age of Internet and Virtual Reality}
\author{Florentin Neumann\thanks{Corresponding author,
\texttt{fneumann@mail.upb.de}\quad
The present work is based on the first author's bachelor thesis
\cite{Florentin} prepared under the supervision of the other two.} 
\and Andrea Reichenberger \and 
Martin Ziegler\thanks{Supported by \textsf{DFG} grant \texttt{Zi\,1009/2-1}
and generously hosted by Prof. Karl Svozil.}}
\institute{\vspace*{-2ex}University of Paderborn, University of Bochum, 
and Vienna University of Technology}
\makeatletter
\renewcommand\maketitle{\newpage
  \refstepcounter{chapter}%
  \stepcounter{section}%
  \setcounter{section}{0}%
  \setcounter{subsection}{0}%
  \setcounter{figure}{0}
  \setcounter{table}{0}
  \setcounter{equation}{0}
  \setcounter{footnote}{0}%
  \begingroup
    \parindent=\z@
    \renewcommand\thefootnote{\@fnsymbol\c@footnote}%
    \if@twocolumn
      \ifnum \col@number=\@ne
        \@maketitle
      \else
        \twocolumn[\@maketitle]%
      \fi
    \else
      \newpage
      \global\@topnum\z@   
      \@maketitle
    \fi
    \thispagestyle{empty}\@thanks
    \def\\{\unskip\ \ignorespaces}\def\inst##1{\unskip{}}%
    \def\thanks##1{\unskip{}}\def\fnmsep{\unskip}%
    \instindent=\hsize
    \advance\instindent by-\headlineindent
    \if@runhead
       \if!\the\titlerunning!\else
         \edef\@title{\the\titlerunning}%
       \fi
       \global\setbox\titrun=\hbox{\small\rm\unboldmath\ignorespaces\@title}%
       \ifdim\wd\titrun>\instindent
          \typeout{Title too long for running head. Please supply}%
          \typeout{a shorter form with \string\titlerunning\space prior to
                   \string\maketitle}%
          \global\setbox\titrun=\hbox{\small\rm
          Title Suppressed Due to Excessive Length}%
       \fi
       \xdef\@title{\copy\titrun}%
    \fi
    \if!\the\tocauthor!\relax
      {\def\and{\noexpand\protect\noexpand\and}%
      \protected@xdef\toc@uthor{\@author}}%
    \else
      \def\\{\noexpand\protect\noexpand\newline}%
      \protected@xdef\scratch{\the\tocauthor}%
      \protected@xdef\toc@uthor{\scratch}%
    \fi
    \if@runhead
       \if!\the\authorrunning!
         \value{@inst}=\value{@auth}%
         \setcounter{@auth}{1}%
       \else
         \edef\@author{\the\authorrunning}%
       \fi
       \global\setbox\authrun=\hbox{\small\unboldmath\@author\unskip}%
       \ifdim\wd\authrun>\instindent
          \typeout{Names of authors too long for running head. Please supply}%
          \typeout{a shorter form with \string\authorrunning\space prior to
                   \string\maketitle}%
          \global\setbox\authrun=\hbox{\small\rm
          Authors Suppressed Due to Excessive Length}%
       \fi
       \xdef\@author{\copy\authrun}%
       \markboth{\@author}{\@title}%
     \fi
  \endgroup
  \setcounter{footnote}{\fnnstart}%
  \clearheadinfo}
\makeatother

\maketitle\vspace*{-5ex}
\begin{abstract}
Inspired by Hofstadter's \emph{Coffee-House Conversation} (1982)
and by the science fiction short story \emph{SAM} by Schattschneider (1988),
we propose and discuss criteria for non-mechanical intelligence.
Firstly, we emphasize the practical need for such tests in view of
massively \mbox{multiuser} online role-playing games (MMORPGs)
and virtual reality systems like \aname{Second Life}.
Secondly, we demonstrate Second Life as a useful
framework for implementing (some iterations of) that test.
\end{abstract}
\begin{minipage}[c]{0.98\textwidth}
\renewcommand{\contentsname}{}\vspace*{-12ex}\small%
\tableofcontents
\bigskip
\end{minipage}
\pagebreak
\setcounter{page}{0}
\setcounter{footnote}{0}
\renewcommand{\thefootnote}{\alph{footnote}}
\newcommand{\footnoteref}[1]{$^\text{\ref{#1}}$}
\section{The Turing Test: A Challenge} 
Artificial Intelligence is often conceived as aiming at simulating or mimicking
human intelligence. A well-known criterion for the success of this endeavour
goes back to Alan Turing: In \cite{Turing50} he described an
\emph{Imitation Game} in which an interrogator gets into a dialogue
with a contestant solely through a teletypewriter and has to find
out whether the contestant is human or not\footnote{It has been pointed
out \cite{Sterrett} that the original paper's intention may have been 
subtly different; however, we adhere to what has become the standard 
interpretation of the Turing Test.}. Turing forecasts:
\begin{quote}\it
I belive that in about fifty years' time it will be possible to programme
computers, with a storage capacity of about $10^9$, to make them play the
imitation game so well that an average interrogator will not have more
than 70 per cent chance of making the right identification after five
minutes of questioning.
\end{quote}
It is impressive how accurate the first part of this prediction
has turned out valid (PCs with 1Gbit=128MBytes of main memory
became common just around the end of the last millenium); 
whereas the \textsf{Loebner Prize},
announced for a computer to succeed with the second part of the prediction has not been achieved so far.
The present-time world record holder 
\texttt{Elbot}\footnote{\myurl{http://www.elbot.com/}}
having convinced 3 out of 12 interrogators of being human.
Nevertheless, it is considered only a matter of years until
the first system passes the test.

\subsection{Turing(-like) Test for Problem Resolution} \label{s:Dilemma}
The early successes in Artificial Intelligence led researchers in this field to be very optimistic. 
Nevertheless, it is important to realize the wildly discussed problems with AI. For instance,
how and which `human' rights and behavioral
constraints should be applied to human-like `robots'?
Such deeply ethical questions underlie many
famous science fiction stories and movies\footnote{%
In fact the authors are strongly convinced of Philosophy
not just as a historical science 
but as a powerful method 
highly relevant to modern life as a guide
and capable of shaping our future reality
as anticipated in science fiction.}; 
e.g. ``\emph{I, Robot}'' by \person{Isaac Asimov} 
with its \emph{Three Laws of Robotics}, compare also
the movie
``\emph{Bicentennial Man}'' starring Robin Williams;
or ``\emph{Do Androids Dream of Electric Sheep}''
by \person{Philip K. Dick}, turned into the movie
``\emph{Blade Runner}'' starring Harrison Ford. 
The plot of the latter specifically evolves around 
the serious problem of how to \emph{detect} if the
entity faced by an interrogator is mechanical
(and may be `killed', if so).

\subsection{Turing Test and Internet} \label{s:Internet}
A major obstacle against passing the test is the vast amount of
background information that every grown human has collected over 
the years of life; which a computer program, typically running
only for some hours, cannot. Instead, developers try to provide
their candidate system with a data base of `knowledge/experience' 
represented by pre-processed answers to specific topics and keywords.

The largest data base, in this sense, is of course the world-wide web. In particular
we have in mind 
\begin{itemize}
\item online encyclopedias, 
 offering pre-compiled and objective background information 
 on almost any conceivable topic;
\item discussion boards, providing a more casual view
 and a subjective counterpart enabling a more human discussion.
\end{itemize}
Both can be efficiently accessed using popular search engines.

Of course, access to such external information may be considered
cheating (and certainly is, if the program connects to an actual human
say via \texttt{ICQ}, see Section~\ref{s:ICQ}). 
As a matter of fact the Loebner Prize rules quite reasonably:
\begin{quote}\it
No entry will be tested by contest management which does not provide, 
on the transmittal media, all necessary programs, interpreters, etc.
\end{quote}

\section{Problems with Mechanical Avatars in the Age of Internet} \label{s:Online}
Section~\ref{s:Dilemma} recalled putative and philosophical problems 
arising with artificial agents in the future;
in fact a not-too-near\footnote{\emph{Blade Runner} 
for instance, produced in 1982, is set in the year 2019} future
of material robots roaming our physical reality.
In the present section we point out that 
the virtual reality of the internet has already turned
the problem of (automatically)
recognizing mechanical\footnote{We avoid the term 
`\emph{Artificial Intelligence}' because of its
philosophical ambiguities and in order to include
the example in Section~\ref{s:GoldFarming}.} avatars 
into a strongly present and practical one. We give four examples for this development.

\subsection{Chatterbots, Spam, and Instant Messengers} \label{s:ICQ}
In 1964--1966, \person{Joseph Weizenbaum} devised a computer program
to parody ``the responses of a non-directional psychotherapist in an 
initial psychiatric interview''. In spite of its technical simplicity
(and to the surprise of its creator)
this turned out to fool many `patients'. A series of successor
software systems and improvements have followed, 
so-called \textsf{chatterbots}, enlargening the 
vocabulary and field of expertise like e.g. the \emph{Artificial Linguistic Internet Computer Entity}
\texttt{A.L.I.C.E.}\footnote{\myurl{http://alicebot.blogspot.com/}}.
In fact, this is the setting for the annual Loebner Prize
in which the jury is presented a candidate system through a terminal
with the explicit goal to determine whether it is human or not.

Infamous spam is probably known to anyone with a computer account:
unsolicited electronic messages
offering cheap\footnote{although true reward is well-known to require previous effort}
satisfaction to deep human desires 
(physical appearance, recognition, sex, money) 
as bait to the financial advantage of the original sender.
The number of such emails literally flooding the internet
raises the problem of automatically and reliably detecting and
deleting (or at least marking) them as an assistance
to the account owner who is otherwise in danger
(or kept exceedingly busy) of missing the `true/important'
messages hidden between spam.
Such tests, however, differ notably from Turing's Imitation Game
in which the underlying communication is immanently one-directional:
The recipient of a putative spam email should not even try to reply
in order to straighten out his/her suspicion, because that will most likely
increase the black market `rating' of his/her email address as `\emph{active}'
and result in even more spam sent to it.

Nevertheless, instant messenger protocols, services, and clients
like \texttt{ICQ}, \texttt{Jabber}, or \texttt{AIM}
nicely complement email as a means
of electronic communication: They are designed for 
immediate, low-overhead and informal exchange of
relatively short messages and thus support a more
interactive and sketchy (and volatile) form of communication, 
rather like remote dialogues than proper letters.
The first author, for instance, can be reached at 
\texttt{ICQ \#232634449}---or 
is that a chatterbot having replaced him?
Indeed, many messenger clients provide 
plugins\footnote{\texttt{ALICE} for instance can be 
merged with \texttt{Miranda}, see
\myurl{http://addons.miranda-im.org/details.php?id=326}.}
for chatterbots to jump in, if the human user is unavailable.
Some messenger accounts are even dedicated bots,
for example \texttt{ICQ \#361718479} (primarily in German).

Presently, we are encountering a trend to synthesize (email-) spam
and instant messenging to so-called \emph{malicious chatterbots}: 
automated electronic advertisement, promotion, and luring
with a new degree of interactivity that email is lacking
and thus raising a very practical urge to detect and quell them
in order to protect the user from such nuissance and danger.
Like with email, such a detector should preferably work mechanically
(i.e. no human interrogator), but unlike email, 
the aspect of interactivity in instant messenging
prohibits any form of offline filter.

\subsection{Game Bots} \label{s:GameBot}
With the first computers (like the \texttt{PDP}), 
first computer games (like \textsf{Spacewar!}) followed soon.
Since then, both computer technology and computer game design
have evolved coherently. Presently the leads have even
switched, video gaming graphics hardware being recognized
as benefitial and used for scientific computing and number
crunching with \textsf{NVIDIA CUDA} and \textsf{ATI Firestream}.
Text-based computer role-playing games date back to the 1974 
\textsf{Dungeons\&Dragons} but were taken to an entire new level
with the above advent of graphics capabilities.
And finally throwing in the internet has resulted 
in the presently immense popularity of 
\textsf{Massively Multiplayer \emph{Online} Role-Playing Games}
(MMORPGs). 
These generally provide the user with a choice of goals and
challenges of various degrees of difficulty. 
The more advanced ones require special \textsf{items}
(virtual tools, weapons, skills, money) that can
gradually be acquired and traded in-game;
see e.g. Figure~\ref{f:DiabloII} in the appendix.

Now some players lack the patience or time of first
passing through all the initial levels---and have their
client mechanically repeat a more or less simple sequence
of movements, each raising only a little amount of 
money\footnote{\label{f:Grind}The repetitive process of 
earning virtual wealth is called \emph{grinding}, 
see \myurl{http://en.wikipedia.org/wiki/Grind\_(gaming)}.}
but over the coarse of virtual time (e.g. one real night)
aggregates enough to cut short the intended game play 
and simply buy the
desired item. Such a client extension is called a
\textsf{Game Bot} and well-known for many MMORPGs
such as \textsf{Diablo II}, \textsf{EverQuest}, 
\textsf{Lord of the Rings Online}, \textsf{Ultima Online}
etc---and an obvious thorn in the side of the
MMORPG's operator who would very much like to
(automatically) identify and, say, temporarily suspend
such a user's avatar.

\subsection{Gold Farming} \label{s:GoldFarming}
As an immediate measure against Game Bots as described above,
modern MMORPGs have included a source of randomness into
their game play. For instance, previously, 
some server-controlled `monster' carrying a minor coin may
re-appear reproducibly at a certain definite position after 
the user has killed it and temporarily left the place,
so that the bot can simply repeat the previous sequence of
moves in order to aggregate wealth; whereas now the monster 
would perhaps re-appear somewhere else, thus requiring 
more advanced and adaptive user interaction.
However, MMORPGs' (simplified) virtual 
economies\footnote{\myurl{http://ezinearticles.com/?id=1534647}}
have spurred real-life counterparts:
`Items' (must not, but) can be purchased for actual money,
e.g. on \texttt{eBay}.
The wage differential and globalization have made 
this a lucrative form of business: computer kids,
mostly in China and South Korea but also elsewhere,
take over the role of the (insufficiently intelligent)
Game Bots and perform (still mostly mechanical and repetitive)
moves with their own avatar to gain virtual wealth
and then sell it for real money on eBay to (mostly western)
players. 

This influence of the floating real world market
to the (basically) fixed exchange rate system
in the relatively small virtual world, has obviously a
considerable impact on its economy.\footnote{\label{f:Economy}With worldwide over 10 million
subscribers in \textsf{World of Warcraft} converse effects are also 
noticable, see \myurl{http://news.cnet.com/2030-1069\_3-5905390.html}.} Again, MMORPGs' operators are therefore
faced with the challenge of identifying and suspending
certain accounts: this time not computer-controlled ones
but those operated by ordinary humans who just happen to 
violate some in-game laws (formally and in the real world: 
end-user licence agreement or terms of service), 
typically from Asia and, being kids, 
often do not speak/write proper English.
Such abuse has led regular \textsf{World of Warcraft} users to
involve\footnote{\label{f:Chinese}%
This is, of course, highly controversial.
It may be argued that in-game laws should be enforced by
an in-game police and not by snitches. Also, the requirement 
of fluent English is dangerously close to racism.
On the other hand the goal of the original Imitation Game
was to distinguish a male from a female opposite,
yet Turing is above any suspicion of sexism.} 
`suspicious' avatars into an online game chat:
definitely a variant of the Turing Test!

\subsection{Non-Player Characters in Second Life} \label{s:SecondLife}
MMORPGs traditionally feature a strict distinction between
player characters (i.e. the real user's virtual counterpart)
and non-player characters (NPCs, e.g. the monstor 
mentioned in Section~\ref{s:GoldFarming}):
the first must be controlled directly and interactively by a human,
choosing from a \emph{pre-}programmed selection of activities
(move, fight, trade, chat, etc.);
the latter are operated by the MMORPG server to exhibit
a level of \emph{fixed-}programmed artificial intelligence.

\begin{figure}[h!]
\centerline{\includegraphics[width=0.86\textwidth]{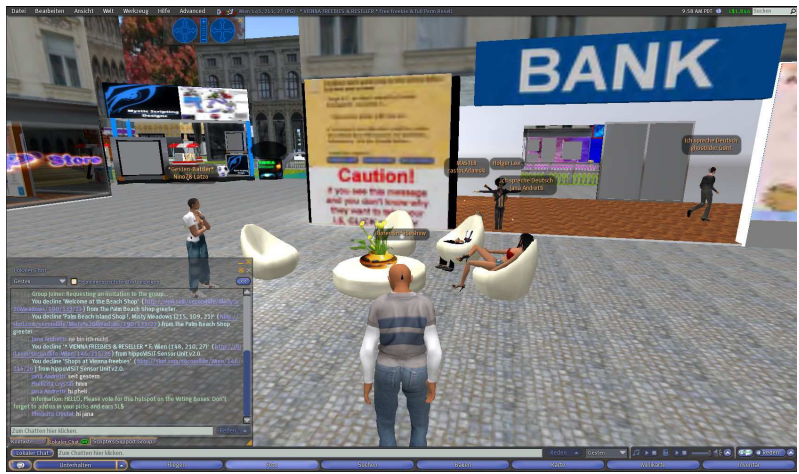}}
\caption{\label{f:SL}Second Life in user view}
\end{figure}

Such a distinction is largely removed in the 
online virtual reality system \textsf{Second Life}.
Here, NPCs are entirely missing and so is, at least 
initially, any form of detailed scenery.
Instead users may freely construct
items and buildings by themselves. Further, they are encouraged 
to place and even to sell them on purchased virtual estate; see Figure~\ref{f:SL}.
Moreover, in striking contrast to MMORPGs,
these objects can be \emph{user-}programmed 
to perform actions on their own.
We demonstrate in Section~\ref{s:TuringComplete} 
that the principal processing capabilities of such a user-created 
object in Second Life coincide with those of a Turing machine and may therefore
be considered the strongest conceivable form of non-human intelligence.

When facing another avatar in Second Life, 
it is therefore not clear whether this 
constitutes indeed another user's virtual representation
or rather a literally `animated' object;
and we consider the problem of distinguishing the latter two cases
as another variant of the Turing Test.

\subsection{Summary and Classification} \label{s:Classification}
The original purpose of the Turing Test was setting a well-defined goal for the
development of artificial intelligence. After 50 years a reconsideration
is advisable. Specifically, the above examples suggest to reverse the focus: 
from devising a mechanical system to pass the test,
towards devising test variants that reliably do distinguish
(at least certain) mechanical systems from (other) human ones.
Depending on the purpose and range of application,
these variants of the Turing Test may 
\begin{enumerate}
\item[i)] 
  employ either a human interrogator or an automated mechanical one.
\item[ii)] 
  distinguish either human from non-human contestants,
  or different classes\footnoteref{f:Chinese}
  of human contestants; or even different classes of non-human ones.
\item[iii)] 
  proceed either passively/observant, or actively challenge the candidate.
\item[iv)] 
  either be restricted to communication via teletype,
  or include more channels of perception provided
  by the virtual reality system under consideration
  such as observing the contestant's movements in virtual space (Gold Farming),
  and perhaps even additional counterparts to human senses
  like hearing / sound of voice.
\item[v)] 
 either permit or prohibit the contestant's online access to the internet.
\end{enumerate}

\section{Variations of the Turing Test}
In Section~\ref{s:Online} new challenges and applications
which are not covered by the original Turing Test have been discussed.
We report now on some known aspects as well as on some new deficiencies of this test. Then we
discuss the Hofstadter-Turing Test as
a stronger variant that mends some of them by taking into
consideration more aspects of human intelligence. Finally,
we prove the weaker Chomsky-Turing Test
undecidable to a deterministic interrogator.

\subsection{Deficiencies of the Original Test}
The Turing Test can be seen as having initiated, 
or at least spurred, Artificial Intelligence as a science.
Nevertheless (or maybe rather: therefore) it is subject
to various criticism and objections,
some raised by Turing himself; cf. e.g. \cite{OppyDowe,Saygin}.

\subsubsection[Restricted Interaction and the Total Turing Test]{%
Restricted Interaction and the Total Turing Test:}
Turing used teletypewriters, a technology of his time,
as a means to hide to the interrogator 
the obviously non-human appearance of 
the computer hardware. He deliberately restricted 
communication and interaction in his test,
so-to-speak \emph{projecting} attention
solely to the intellectual (if not platonic)
features of the contestant.
In science fiction 
movies on the other hand, 
physical appearance also plays an important role;
recall Section~\ref{s:Dilemma}.
More precisely, Turing originally ignores
most human senses entirely---channels of perception 
which in the 1950ies were indeed unforseeable 
to artificial simulation. 

However, this situation has changed dramatically
with the rise of multimedia and virtual reality systems
(recall Section~\ref{s:SecondLife})
including not just 3D vision and directional sound
but even haptic feedback \cite{Haptic}, yielding a rather
drastic, undescribable experience of actual reality.
In order to take this into account,
the so-called \emph{Total} Turing Test has been proposed
as an advanced goal of Artificial Intelligence \cite{Harnad}.

\subsubsection[Physical Reality and Threat]{Physical Reality and Threat:}
Extending the above reproaches, one may argue that mere
interaction with the interrogator---even if sensual---does
not suffice to classify as intelligent. 
A considerable part of human (as opposed to animal)
existence arises from, and evolves around constructing new objects, tools, and weapons 
within and from its natural physical environment as a means to meet with
everyday challenges and threats. In fact, according to Darwin,
such threats are the origin of and the catalyst to the degree
of learning and creativity we consider typically human.

This is an aspect not covered even by the
Total Turing Test---but well present in the
Hofstadter-Turing Test described in Section~\ref{s:Hofstadter} below.

\subsubsection{Anthropocentrism} \label{s:Anthropocentrism}
is in our opinion the most serious deficiency,
in fact of Artificial Intelligence as a whole:
it starts with the (usually implicit) hypothesis
that humans \emph{are} intelligent (``cogito ergo sum''),
and then proceeds in \emph{defining} criteria for intelligence
based on resemblance to various aspects of humans.
Voluntarily equipping oneself with a blind spot
seems like a strongest disproof against the initial hypothesis.
On a less sarcastic level, anthropocentrism is known
to have caused many dangerous and long-lasting errors
throughout human history.
Of course we have no simple solution to offer either
\cite[p.509]{Saygin},
other than to constantly remain open and alert
against such fallacies.

\subsection{Hofstadter-Turing Test} \label{s:Hofstadter}
In 1988, \person{Dr. Peter Schattschneider} published a science fiction short story
``\emph{SAM}'' \cite{Schattschneider} in the series of the computer magazine \emph{\ct}.
It begins from the point of view of an unspecified `being'
finding itself wandering an increasingly complex environment,
later revealed to be controlled by a `programmer',
and eventually arriving at a computer terminal where
it starts setting up a similar virtual environment and wanderer,
thus passing what is revealed (but not specified any further)
as the \emph{Hofstadter Test}:
\begin{quote}\it
Im Hofstadter-Test wird das Programm mit einer Krisensituation konfrontiert, 
in der es st\"{a}ndig gezwungen ist, seine Lage zu \"{u}berpr\"{u}fen,
um \"{u}berleben zu k\"{o}nnen. Der Hofstadter-Test gipfelt in der Forderung,
ein intelligentes, bewu\ss{}tes Programm zu erstellen.
\end{quote}
This story may have been inspired by 
\person{Douglas R. Hofstadter}'s \emph{Coffee-House Conversation} \cite{Hofstadter}
of three students, Chris (physics), Pat (biology), and Sandy (philosophy)
ending with the following lines:
\begin{quote} \it
\person{Chris:} 
If you could ask a computer just one question 
in the Turing Test, what would it be?

\person{Sandy:} Uhmm\ldots

\person{Pat:} How about this: 
``\emph{If you could ask a computer just one question in the Turing Test, what would it be?}''
\end{quote}
Observe the recursive self-reference underlying both, 
this last question and Schattschneider's story ``\emph{SAM}''
as well as Turing's famous article \cite{Turing36}
proving by diagonalization and self-reference
that the question of whether a Turing machine eventually terminates
(i.e. the Halting problem, cf. e.g. Equation~\ref{e:Halting} on
page~\pageref{e:Halting}) is undecidable to a Turing machine.
Picking up \cite{Schattschneider}, we arrive at the following:

\begin{definition}[Hofstadter-Turing Test] \label{d:Hofstadter}
For an entity to pass the Hofstadter-Turing Test means to devise
\begin{enumerate}
\item[i)] a virtual counterpart resembling its own environment ~ and
\item[ii)] a computer program which succeeds in
  recognizing itself as an entity within this virtual environment and
\item[iii)] in turn passes the Hofstadter-Turing Test.
\end{enumerate}
\end{definition}

\subsubsection[On the Self-Reference]{On the Self-Reference:}
\label{s:Recursion}
Because of Condition~iii), Definition~\ref{d:Hofstadter}
is in danger of being circular. We want to address this
important issue in three different ways.

\paragraph{In classical logic} the problem can be removed
by `unfolding' the condition in requiring the existence
of a countably infinite sequence of virtual environments
and entities such that the $(n+1)$-st are created
by and within the $n$-th.
This does not provide any way of operationally performing
this test but at least makes the definition valid.

\paragraph{In practice and pragmatically,} 
once the first few\footnote{\label{f:Implement}%
Section~\ref{s:Implement} shows that
at least the first author has passed the
initial $2.5$ levels.}
levels $n$ have succeeded in creating
their successor $n+1$, one would likely be content to
abort any further recursion
and state with sufficient conviction that the 
initial entity has passed the test.

\paragraph{In his short story \person{Schattschneider}}
gave another resort to the infinite re-iteration:
\begin{quote}\it
Wenn der Spieler gewinnt, gibt es \emph{keine}
Wiederholungen. Bewu\ss{}te Programme sind so
verschieden wie du und ich. Sie k\"{o}nnen
dem Kreislauf entrinnen. [\ldots]
Sein Ziel w\"{u}rde erreicht sein,
wenn er ein Programm geschrieben hatte,
das in der Lage war, sich selbst zu erkennen.
\end{quote}
We shall return to this remark in Section~\ref{s:Levels}.

\subsubsection[Critical Account]{Critical Account:}
We readily admit that the Hofstadter-Turing Test
does not provide the ultimate solution
to the problems of Artificial Intelligence
and mention three reproaches.

\paragraph{Anthropocentrism} is, again, present in its strongest form
by requiring our human physical world to be the first
and thus modelled by all iterated virtual counterparts
according to Condition~i) environments.

In fact it seems that the common conception of a `virtual environment'
is highly biased and restricted. Even a critic of Platonic realism
will find it hard to explain why a computer's 
`digital world' of \texttt{0}s and \texttt{1}s should 
\emph{not} be considered an ontological reality 
but be required to reflect 
what humans consider (!) as real.
Even more, questions of intelligence and consciousness 
are irrelevant within the abstract `world' of programs;
they only arise through the sociocultural interface
of virtual role-playing systems.

\paragraph{The Problem of Other Minds}\footnote{\label{f:Otherminds}The 
\textsf{Problem of Other Minds} raises following issue:
Given that I can only observe the behaviour of others, 
how can I know that others have minds? And this issue
particularly applies to the question of how
to detect 'mechanical minds' \cite{Tetens}.
An answer to the latter seems at least as out of range
as one to the former.
} is a well-known philosophical issue which arises here, too:
Is the contestant (sequence!) required to exhibit
and visualize the virtual environment he/she has created?
Can we even comprehend it, in case it is a purely digital one?
How about patients with locked-in syndrome:
does the single direction of their communication capabilities
disqualify them from being intelligent?
In final consequence, one arrives at the well-known problems
of Behaviorism. 

\paragraph{``Humans are intelligent''} 
used to be the first axiom of any test for true intelligence.
But is actually any person able to pass 
(pragmatically at least some initial levels of)
the Hofstadter-Turing Test?
Recall he has to succeed\footnoteref{f:Implement}
in creating a virtual entity of true intelligence.

\subsection{Chomsky-Turing Test} \label{s:Chomsky}
As pointed out, several applications prefer 
an automated form of Turing-like tests (recall 
Item~i) in Section~\ref{s:Classification}).
The present section reveals that this is, 
unfortunately, infeasible in a strong sense.
Referring to the Theory of Computation 
we formally prove that even powerful oracle Turing machines
(capable of solving e.g. the Halting problem)
cannot distinguish a human contestant from
the simplest (abstract model of a) computing device
in Chomsky's hierarchy, namely from a finite automaton.

\subsubsection[Reminder of the Theory of Computation]{Reminder of the Theory of Computation:}
In his PhD thesis \cite{TuringDiss} Turing considered an extension
of `his' 1936 machine which he denoted as \emph{o-machine}. 
This is nowadays known as \textsf{oracle Turing machine} and permitted
during its computation to repeatedly submit a number (or binary string)
$x$ it may have calculated so-far to some hypothetical external
device called oracle and formalized as a set $\calO\subseteq\IN$
(or $\calO\subseteq\{0,1\}^*$). This device will then, and
deterministically, provide within one step an either positive
``$x\in\calO$'' or negative answer ``$x\not\in\calO$'' for the
Turing machine to rely on in the subsequent steps of its calculation.
Depending on the choice of oracle, such a machine can be very powerful; 
for instance for $\calO:=H$ it can decide the (otherwise undecidable)
Halting problem
\begin{equation} \label{e:Halting}
H \quad=\quad \{\langle M\rangle: \text{Turing machine
$M$ terminates on the empty input}\}
\end{equation}
simply by passing the encoded input machine $\langle M\rangle\in\IN$, 
whose termination is under question, right on to the oracle.
On the other hand, even a machine with oracle access to $H$
provably cannot decide the so-called relativizes Halting problem
$H^H$, that is the question of whether another given
machine with oracle access to $H$ terminates on the empty input or not.
Iterating, one arives at \person{Stephen C. Kleene}'s infinite 
(in fact transfinite) hierarchy of oracle machines of strictly 
increasing computational power.

\medskip
In the converse direction, namely concerning models of computation weaker
than the Turing machine, \person{Noam Chomsky} had devised a
four-level hierarchy for classifying (originally natural, nowadays
usually formal) languages. Among the big successes of
Theoretical Computer Science (e.g. Turing Awards for 
\person{Michael Rabin} and \person{Dana Scott}) was an
equivalent characterization of each level, 
in terms of grammars generating the languages therein
as well as in terms of machine models accepting these languages:

\begin{center}
\begin{tabular}{c||c|l}
\# & grammars & machines \\
\hline 
0 & unrestricted & Turing machines \\
1 & context-sensitive & linearly space-bounded nondeterministic Turing machines\\
2 & context-free & pushdown automata \\
3 & regular & finite state machines
\end{tabular}
\end{center}
Finite state machines are generally considered as a model for 
simple control units like those of digital watches, elevators,
or washing machines. In fact the \textsf{Pumping Lemma} reveals
them as too weak to merely check a boiled down variant of
syntactical correctness of a mathematical formula, namely
the question of whether a given string $x$ contains
as many opening brackets (or \texttt{0}s) as closing ones (or \texttt{1}s).

Chomsky's Hierarchy originally pertains to languages,
that is to sets $L\subseteq\{0,1\}^*$ of finite binary strings
and their associated word problems of deciding,
given some $x\in\{0,1\}^*$, whether $x\in L$ holds or not;
input $x$, output \texttt{yes/no}.
However, the above machine models have natural
and well-established extensions for dialogue-like problems.
For instance, a Turing machine may `request' further user input to be typed onto its tape 
by printing some special \emph{prompt} symbol;
and finite state machines become \textsf{transducers}
such as \emph{Moore machines} or, equivalently, \emph{Mealy machines}.

\subsubsection[Strong Undecidability of the Chomsky-Turing Test]{%
Strong Undecidability of the Chomsky-Turing Test:}
As mentioned above, finite state machines and transducers (Chomsky level 3)
are models of computation with immensely limited capabilities.
Turing machines are located at the other end of the hierarchy (level 0).
If we wish to include non-mechanical language processors,
humans could be defined to reside at level -1.
In which case the goal of (Artificial Intelligence and) the Turing Test amounts 
to (separating and) distinguishing level 0 from level -1.

The present section goes for a much more modest aim:
\begin{definition}[Chomsky-Turing Test] \label{d:Chomsky}
The goal of the Chomsky-Turing Test is to distinguish
Chomsky level 3 from level -1.
\end{definition}
We establish that such a test cannot be performed mechanically.
A first result in this direction is a well-known consequence of
\textsf{Rice's Theorem} in computability theory;
cf. e.g. \mycite{Theorem~5.3}{Sipser}.

\begin{fact} \label{f:Rice}
The language \textsc{RegularTM}, defined as
\[ 
\big\{ \langle M\rangle : \text{the language $L(M)\subseteq\{0,1\}^*$
accepted by Turing machine $M$ is regular} \big \} \]
is undecidable to any Turing machine.
\end{fact}
It is, however, decidable by an appropriate oracle machine,
namely taking \textsc{RegularTM} itself as the oracle.
Moreover, the above mathematical claim does not quite apply to the setting
we are interested in: It supposes the encoding (G\"{o}del index)
of a contestant Turing machine $M$ to be given and to decide whether
$M$ acts as simple as (but of course not is) a finite state machine;
whereas in Turing-like tests, the contestant may be human
and is accessible only via dialogue. 
Such a dialogue amounts to a sequence 
$(x_1,y_1,x_2,y_2,\ldots,x_n,y_n,\ldots)$ of finite strings $x_i$
entered by the interrogator and answered by the contestant
in form of another finite string $y_i$ upon which the
interrogator adaptively enters $x_{i+1}$, the
contestant replies $y_{i+1}$, and so on round by round.
For this setting, we have the following replacement to
Fact~\ref{f:Rice}:

\begin{proposition} \label{p:ChomskyTuring}\text{ }
\begin{enumerate}
\item[i)] It is impossible for any deterministic interrogator
to recognize with certainty and within any finite number of 
communication rounds $(x_1,y_1,x_2,y_2,\ldots,x_n,y_n)$
that the answers $x_i$ provided by the contestant 
arise from a transducer (Chomsky level $=3$).
\item[ii)] It is equally impossible
in the same sense to recognize with certainty 
that the answers arise from a device on 
any Chomsky level $<3$.
\end{enumerate}
\end{proposition}
Thus, we have two separate (negative) claims, corresponding
to (lack of) both recognizability and co-recognizability 
\mycite{Theorem~4.16}{Sipser}.
More precisely, the first part only requires the interrogator 
to report ``\texttt{level} $=3$'' within a finite number of
communication rounds in case that the contestant is a
transducer but permits the dialogue to go on forever,
in case it is another device; similarly for the second part.
Also, the condition of a deterministic interrogator is 
satisfied even by oracle Turing machines. Hence, this can be called a strong form of undecidability result.

\begin{proof}[Proposition~\ref{p:ChomskyTuring}]
Both claims are proven indirectly by `tricking' a putative
interrogator $I$. For i) we first face $I$ with some
transducer $T$; upon which arises by hypothesis a finite
dialogue $(x_1,y_1,\ldots,x_n,y_n)$ ending in $I$
declaring the contestant to be a transducer (``\texttt{level} $=3$'').
Now this transducer $T$ can be simulated on any lower 
Chomsky level by an appropriate device $D$ 
exhibiting an input/output behavior identical to $T$. 
Since $I$ was supposed to
behave deterministically, the very same dialogue 
$(x_1,y_1,\ldots,x_n,y_n)$ will arise when presenting 
to $I$ the contestant $D$---and end in $I$ 
erroneously declaring it to be a transducer.
\\
The proof of Claim~ii) proceeds similarly: first present
to $I$ some device $D$; which by hypothesis leads to
a finite dialogue $(x_1,y_1,\ldots,x_n,y_n)$ and the
report ``\texttt{level} $<3$''. Now it is generally impossible 
to simulate $D$ on a transducer since Chomsky's Hierarchy
is strict \cite{Sipser}. However any fixed finite dialogue
can be hard-coded into some transducer $T$; and 
repeating the interrogation with this $T$ results by
determinism of $I$ in the same, but now wrong,
answer ``\texttt{level} $<3$''.
\qed\end{proof}
Observe that the above proof of Proposition~\ref{p:ChomskyTuring}ii)
involves transducers $T$ of unbounded size in order to hard-code 
the fixed but arbitrary dialogue. In fact if an upper bound on
the number of states of $T$ is given, the problem does become decidable
and turns into a well-studied topic within the field of \emph{Model Checking},
cmp. e.g. \mycite{Section~7.1}{Yannakakis}.
\section[Implementing the Hofstadter-Turing Test in Second Life]{%
Implementing the Hofstadter Test in Second Life} \label{s:Implement}
Recall (Section~\ref{s:Hofstadter}, Definition~\ref{d:Hofstadter})
that the goal of the test is to implement a virtual reality system
and an artificial entity therein which in turn passes the Hofstadter
test. Fortunately, there is a variety of virtual reality
systems available, so the first level of the test can be
considered accomplished. However, in order to proceed to the next
level, this system has to be freely programmable on the virtual
level---and to the best of our knowledge, this is presently only
supported by Second Life.

Section~\ref{s:TuringComplete} demonstrates
that the scripting language provided by Second Life
indeed is Turing-complete, that is, a programming environment as
powerful as possible. Further, we succeeded in implementing some few inital levels of
the Hofstadter-Turing Test (Section~\ref{s:Sideshow}) within Second Life .

\subsection{Turing-Completeness of the Linden Scripting Language} \label{s:TuringComplete}
\begin{figure}
\centerline{\includegraphics[width=0.96\textwidth]{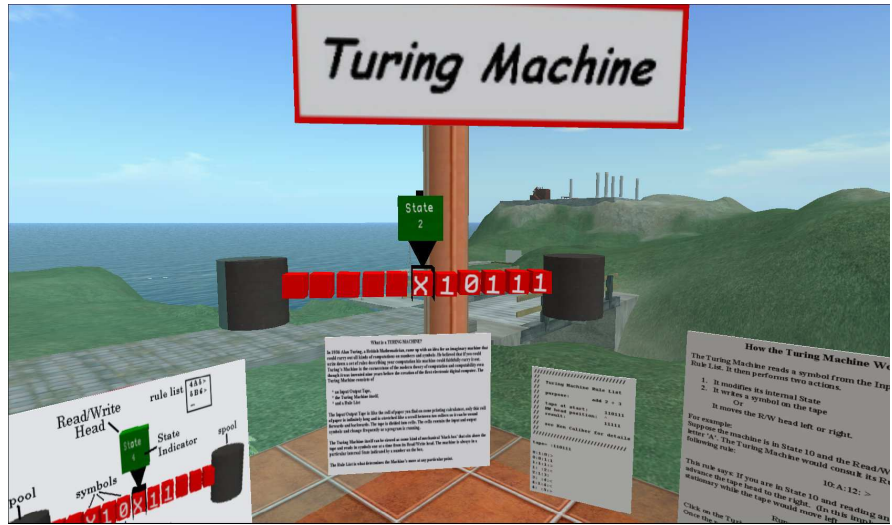}}
\caption{\label{f:SLTM}A virtual Turing machine implemented 
by \person{Prof. Dr. Kenneth Schweller} and students from the
\emph{Buena Vista University, Iowa} in Second Life
at coordinates \myurl{http://slurl.com/secondlife/Buena\%20Vista/87/72/24}}
\end{figure}

Second Life is advertised, among others, for educational purposes.
Correspondingly, 
\person{Dr. Kenneth Schweller} from the \emph{Buena Vista University} (Iowa)
has used it to implement and graphically visualize the operation of 
a Turing machine (see Figure~\ref{f:SLTM}).
This seems to imply that Second Life is Turing-complete.
However, closer investigation reveals some caveats and 
restrictions---which may not (yet) be of practical relevance
but are important from the fundamental point of view of
rigorous computability theory. In fact, the technical backbone
of Second Life (running on a server farm of
the \textsf{Linden Lab} company) raises doubts if it
can actually provide the unlimited computational resources 
required for a truely Turing-complete environment.
Specifically,
the \textsf{Linden Scripting Language} (LSL) is primarily intended to
animate objects in Second Life and each such script is limited
to an overall memory consumption of at most 16kB. Although this may
seem sufficient for most practical purposes arising in Second Life,
a digital computer with constant-size storage cannot be Turing-complete
because the question of its termination 
(as opposed to the Halting problem of a Turing machine)
is decidable to a Turing machine; cf. e.g. \cite[p.178]{Sipser}.
However, this seemingly fundamental restriction can be avoided
using the trick of cascading: A script executed within an object
may initiate an unlimited number of further scripts and send 
messages to them. In this way one can thus implement a linked 
list of unbounded length and linearly accessable via message
passing forth and back: just like a Turing machine's tape.

\subsection{Two-And-Half Iterations of the Hofstadter-Turing Test} \label{s:Sideshow}
\begin{figure}[h!]
\centerline{\includegraphics[width=0.96\textwidth]{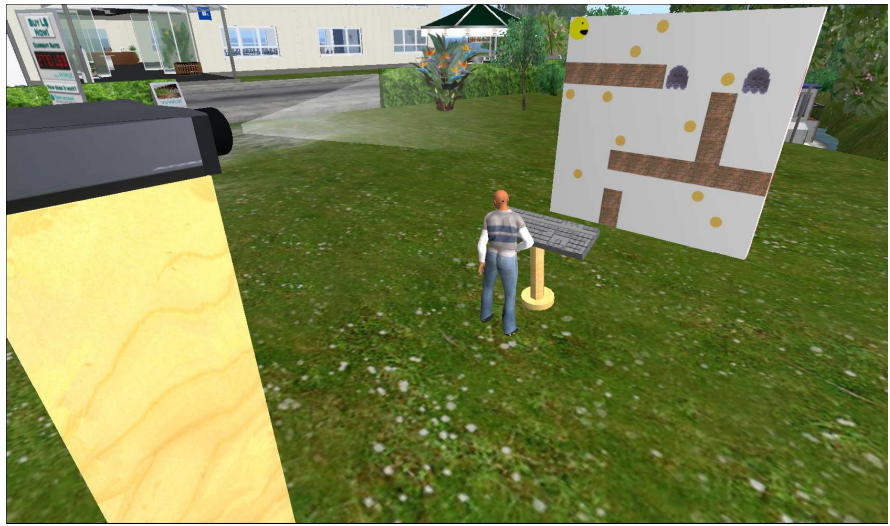}}
\caption{\label{f:SLpacman}An avatar playing a variant 
of game classic \textsf{Pac-Man}
within Second Life available at coordinates
\myurl{http://slurl.com/secondlife/Leiplow/176/136/33}}
\end{figure}

We have succeeded in implementing within Second Life the following virtual scenario:
a keyboard, a projector, and a display screen. 
An avatar may use the keyboard to start and play a 
variant of game classic \textsf{Pac-Man},
i.e. control its movements via arrow keys;
see Figure~\ref{f:SLpacman}.
(For implementation details, please refer to \mycite{Section~4}{Florentin}.)
With some generosity, this may be considered as 
$2.5$ levels of the Hofstadter-Turing Test:

\begin{description}
\item[1st:]
The human user installs Second Life on his computer
and sets up an avatar.%
\item[2nd:]
The avatar implements the game of \textsf{Pac-Man} within Second Life.
\item[3rd:]
Ghosts run through the mace on the virtual screen.
\end{description}
Observe that the ghosts indeed contain some (although 
admittedly very limited) form of intelligence represented by a simple strategy to pursue pacman.

\section{Concluding Remarks}\label{s:Levels}
We have suggested variations of the (standard interpretation of)
the Turing Test for the challanges arising from new technologies such as internet and
virtual reality systems. 
Specifically to the operators of MMORPGs and of Second Life,
the problem of distinguishing mechanical from human-controlled
avatars is of strong interest in order to detect putative abuse.
Indeed, contemporary multimedia technology makes it much easier
for an artificial being to convincingly resemble a human's
virtual counterpart, thus obliviating Turing's original 
restriction of purely teletypewriter-based interaction.
Correspondingly (and in spite of the underlying anthropocentrism)
it seems fair to require a virtual artificial intelligence
to become aware of (and self-aware within) 
its virtual environment; and to employ it:
this leads to the Hofstadter-Turing Test.

\subsection{On Levels of Reality and Their Interaction}
The question of ontology is an old philosophical one: what is real(ity)?
This term, however, has been ``hijacked'' and restricted
in computer science to information representation by means
of data structures and for data exchange.
In the context of virtual reality, though,
one returns to the original meaning:
are the digital worlds of World of Warcraft
and Second Life `real'?
For many of their millions of human users/inhabitants,
they at least constitute a strong surge to spend
large parts of their life online, 
often on the verge of addiction.
There, they meet friends, create homes, 
fight enemies, do commerce etc.
and may choose and alter their appearance,
thus being released from any physical impairments;
exempted even from mortality and the laws of physics!
It seems fair to say that such users \emph{transit}
at least partly to this new reality---which 
by itself is not necessarily bad at all; and which
deserves the real world's verdict of addiction
only because the transition remains incomplete.

An particular feature of the Hofstadter-Turing Test
are the iterated levels of virtual reality it requires to be 
created one within another. Each one of these iterated levels
can be seen as an encapsulated virtual reality, 
transparently contained by the one at the next higher level,
like the skin layers of an onion.
Similarly, behind the visible virtual reality of Second Life
consisting of `idealized' (i.e. platonic)
geometric/architectural objects and avatars,
there lies hidden the invisible and abstract virtual reality
of programs and scripts that control and animate these objects.

This suggests an interesting (new?) approach to the 
philosophical problem of ontology: maybe reality should
generally be considered composed of layers.
In fact such a concept is present in many religions,
although going in the opposite direction:
where Hofstadter-Turing proceeds recursively to inner
and inner layers, Buddhist cosmology for instance promotes 
a hierarchy of `realms' ranging from 
\emph{naraka} up to \emph{Brahma},
from which all lower worlds can be perceived
and as a means to break out from the infinite cycle of rebirth
and to raise on this hierarchy, Buddha
teaches (self-)cognition and enlightment---which
resembles Schattschneider's proposed resort
from the infinite recursion in the Hofstadter-Turing Test,
recall Section~\ref{s:Recursion}.

Less esoterically speaking, the possibility and means to break
through the confinements of one's reality seems interesting
enough. In terms of levels of realities, this includes the
question of whether and how different such levels may interact.
That a higher level can influence
a lower one, should be pretty obvious from the above examples:
the short story \textsf{SAM}, the Hofstadter-Turing Test,
MMORPGs, and Second Life.
But careful reconsideration reveals also
effects in the converse direction:
\begin{itemize}
\item A video game addict socially isolating himself,
  loosing his job and/or health.
\item Virtual items being sold for real money as in Gold Farming;
 cmp. Footnote\footnoteref{f:Economy}.
\end{itemize}
In fact the virtual success of a virtual reality system
is closely tied to its owner's economic situation in real life.
So close that it has resulted in 
\begin{itemize}
 \item actual law-suits for breach of `virtual' laws and unfair trade practices
(see \textsf{Bragg vs. Linden Lab}).
\end{itemize}


\newpage
\begin{appendix}
\section{Screenshots}
\begin{figure}
\centerline{\includegraphics[width=0.86\textwidth]{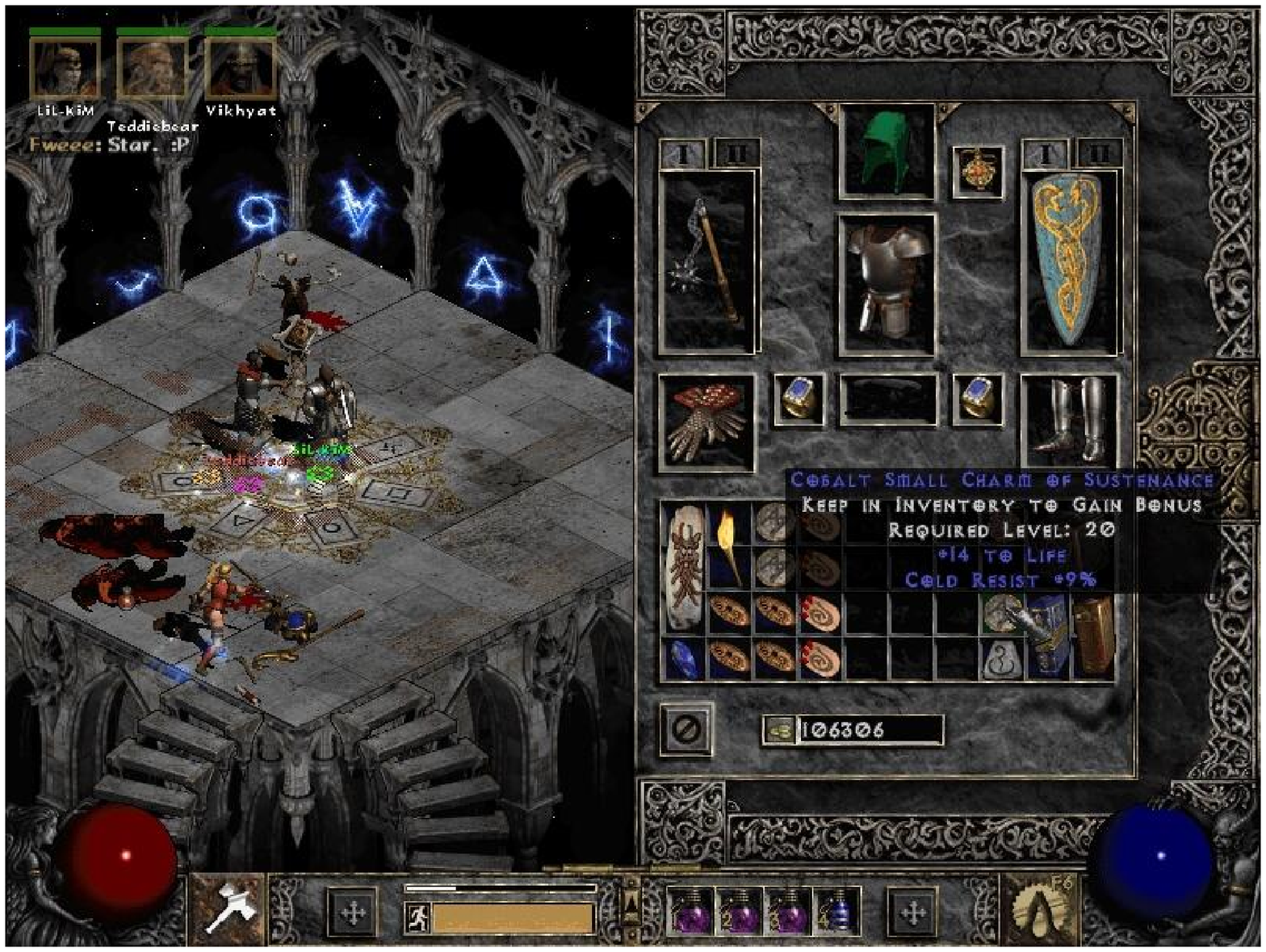}}
\caption{\label{f:DiabloII}Diablo II:
3 players are earning items after a fight, storing them in the players inventory.}
\end{figure}

\end{appendix}
\end{document}